\documentclass[letterpaper, 10 pt, conference]{ieeeconf}  

\IEEEoverridecommandlockouts                              

\usepackage{tabularx}
\usepackage{xcolor}

\usepackage{rotating, multirow}
\usepackage{makecell}

\usepackage[pagebackref,breaklinks,colorlinks]{hyperref}

\title{\LARGE \bf
Detecting Endangered Marine Species in Autonomous Underwater Vehicle Imagery Using Point Annotations and Few-Shot Learning}

\author{Heather Doig$^{1}$, Oscar Pizarro$^{1,2}$, Jacquomo Monk$^{3}$ and Stefan Williams$^{1}$
\thanks{$^{1}$H. Doig, O. Pizarro and S. Williams are with the Australian Centre for Robotics, University of Sydney, Australia. {\tt\small(heather.doig, oscar.pizarro, stefan.williams)@sydney.edu.au}}
\thanks{$^{2}$O. Pizarro is also with the Marine Technology Department, Norwegian University of Science and Technology, Trondheim, Norway.}
\thanks{$^{3}$ J. Monk is with the Institute for Marine and Antarctic Studies, University of Tasmania, Australia. {\tt\small{jacquomo.monk@utas.edu.au}} }
}

\begin{document}

\maketitle
\thispagestyle{empty}
\pagestyle{empty}

\begin{abstract}

One use of Autonomous Underwater Vehicles (AUVs) is the monitoring of habitats associated with threatened, endangered and protected marine species, such as the handfish of Tasmania, Australia.   Seafloor imagery collected by AUVs can be used to identify individuals within their broader habitat context, but the sheer volume of imagery collected can overwhelm efforts to locate rare or cryptic individuals.  Machine learning models can be used to identify the presence of a particular species in images using a trained object detector, but the lack of training examples reduces detection performance, particularly for rare species that may only have a small number of examples in the wild.  In this paper, inspired by recent work in few-shot learning, images and annotations of common marine species are exploited to enhance the ability of the detector to identify rare and cryptic species.  Annotated images of six common marine species are used in two ways.  Firstly, the common species are used in a pre-training step to allow the backbone to create rich features for marine species.  Secondly, a copy-paste operation is used with the common species images to augment the training data.  While annotations for more common marine species are available in public datasets, they are often in point format, which is unsuitable for training an object detector.  A popular semantic segmentation model efficiently generates bounding box annotations for training from the available point annotations.  Our proposed framework is applied to AUV images of handfish, increasing average precision by up to 48\% compared to baseline object detection training.  This approach can be applied to other objects with low numbers of annotations and promises to increase the ability to actively monitor threatened, endangered and protected species.

\end{abstract}

\section{Introduction}
\begin{figure}[!htbp]
    \centering
    \includegraphics[width=0.47\textwidth]{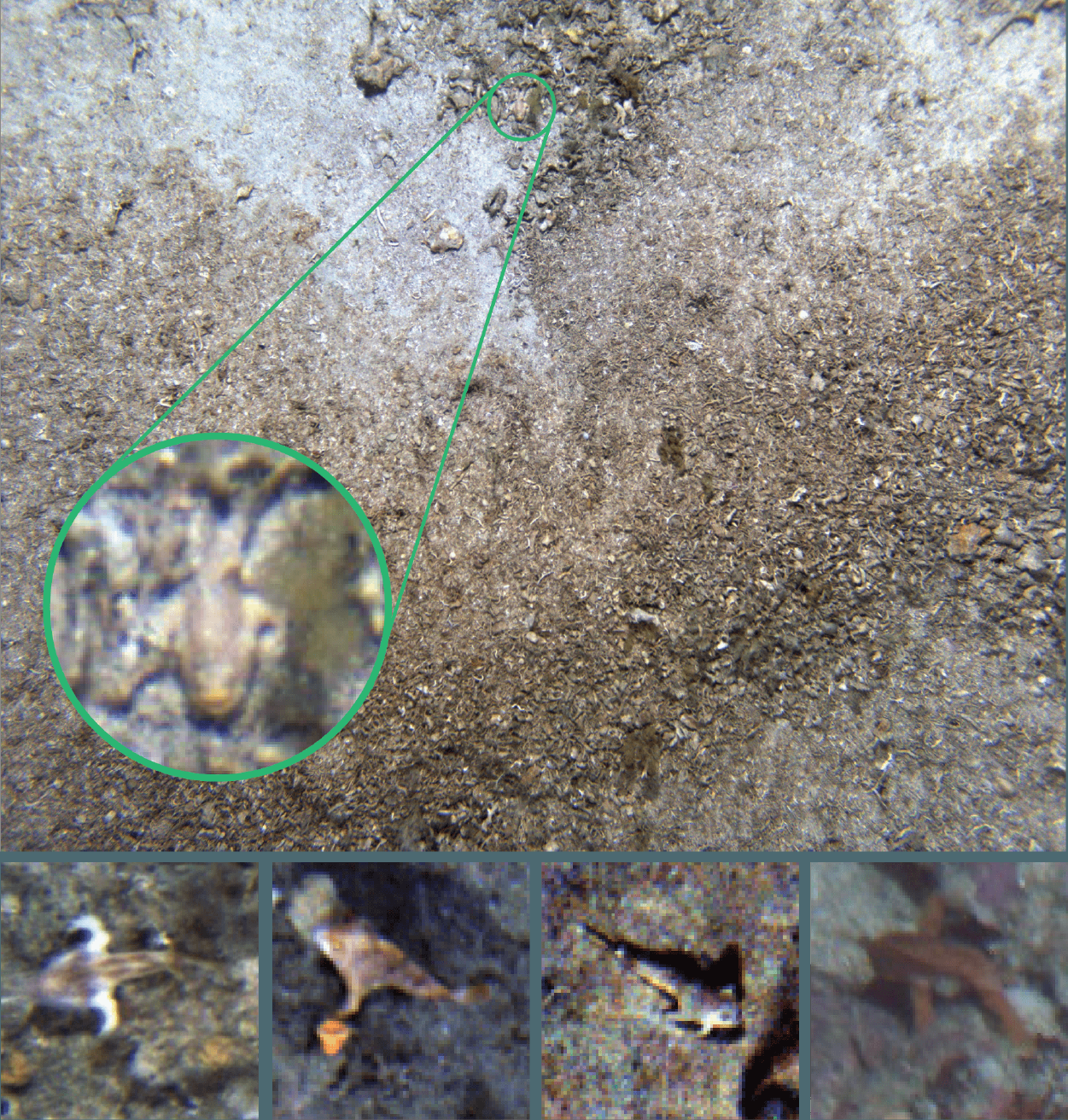}
    \caption{Handfish in imagery captured by \emph{AUV Sirius} and \emph{AUV Nimbus}.  Manually identifying a small and cryptic species like handfish against a complex background is time-consuming and can lead to missed observations. The top image shows a complete image from \emph{AUV Nimbus} with a single handfish. There were 7 images with handfish identified out of 11,171 images captured during the mission.  The bottom row shows cropped examples of handfish.  Images downloaded from IMOS-UMI Squidle+, \href{http://squidle.org}{squidle.org}.}
    \label{fig:handfish}
\end{figure}

Handfish (\textit{Brachionichthyidae}) are a rare and iconic species of bony fish found in southern waters of Australia, with a diversity hotspot for the species around Tasmania. Of the 14 confirmed species of handfish, half of the species are listed as either Endangered or Critically Endangered by the International Union for Conservation of Nature (IUCN), making them the most threatened family of bony fishes in the world ~\cite{Stuart-Smith2020}. One of the species, the red handfish (\textit{Thymichthys politus}), is thought to have less than 100 adults left in the wild. Many of these handfish species are believed to be confined to shallow waters, where impacts from human activities (such as pollution, habitat destruction and terrestrial runoff) and climate change are at their greatest. However, in October 2021, the endangered and very rare pink handfish (\textit{Brachiopsilus dianthus}) was seen for the first time since 1999, in footage from a baited remote underwater video at a depth of 150 m~\cite{Perkins2022Changes}. The discovery has prompted researchers to reconsider the depth range for some of these handfish species, giving optimism that deeper waters may provide some refuge from the impacts causing their population declines.  

The low abundance and elusive nature of threatened species, like handfish, make it challenging to monitor their populations.  Monitoring of handfish populations, like many other shallow water species, has historically been undertaken using SCUBA-diver-based methods and more recently using eDNA~\cite{Perkins2022Changes}. 
The recent detection of listed handfish in deeper waters highlights the need for remote tools to collect high-resolution seafloor imagery such as Autonomous Underwater Vehicles (AUVs). AUVs have been widely used to assess and monitor changes in benthic organisms~\cite{Monk2018Evaluation,Perkins2021AnalysisOfATime,MassotCampos2023}. 

AUV imagery has historically relied on manual annotation to identify organisms in the images. However, this process is slow and time-consuming. Additionally, most handfish species are less than 150mm long and exhibit highly cryptic behaviour, further complicating their detection and identification as shown in Figure~\ref{fig:handfish}. If handfish are present in an AUV mission, they only appear in around 0.2\% of images captured.  A deep-learning object detector could significantly reduce the effort required to identify handfish species within AUV imagery. This approach could also improve data quality, helping gather fundamental population data such as spatial distributions and trends in abundance, critical for the conservation of handfish species.

Object detectors use deep learning models to locate an object with a bounding box and classify the object using a percentage confidence level.  Bounding box annotations with the object class are used to train the detector.  Ideally, the detector is trained with large amounts of data to provide high performance~\cite{Kang2019FewShot,Inoue2018CrossDomain,Zhang2022C2FDA,Munir2023}, but this is often not available for underwater images~\cite{Liu2020Towards,Er2023Research}.  Annotations of marine species are limited due to the time and expertise required but also because the species may be very rare with low populations in the wild~\cite{Bessell2022Prioritising,Bessell2024Population}. For example, the top image of Figure~\ref{fig:handfish} was one of seven images found with handfish (single individual) out of 11,171 images captured on a mission by \emph{AUV Nimbus} in 2023.  

Reduced performance due to domain shift is another issue for deep learning models trained with underwater images~\cite{Langenkamper2020}. 
AUV surveys provide high-quality images taken under consistent operating conditions during a mission, such as altitude, camera, lighting and water conditions.  Images are captured from a birds-eye camera pose at a relatively constant altitude, ensuring that species are shown from similar angles and scales.  While the images are captured with consistent operating parameters in the same mission, images for training and at test time are often from different missions and may potentially be from  different vehicles or imaging systems.  Domain shift can occur where the detector's performance during testing is reduced compared to training due to the differences in the conditions in which the images were captured.  

In this work, we present a framework for detecting rare and cryptic marine species to address the issues of low numbers of annotations and domain shift, inspired by few-shot learning.  This technique trains a model to detect both common base classes with large amounts of examples as well as rare or novel classes with only a few examples~\cite{Kang2019FewShot,Wang2020Frustratingly}.  We use six common marine species as base classes to pre-train the detector before fine-tuning with the novel class of handfish to compensate for the low number of novel class annotations.  

In addition, the annotations of the base classes are used in a copy-paste operation to augment the training data based on~\cite{Ghiasi2021SimpleCopy}.  This operation either copies and pastes the segmented handfish to base class images or copies the segmented base class instances to handfish images.  This two-way operation is randomly applied and aims to reduce domain shift by adding the images with the base classes to training while also addressing the low number of annotations.   

Squidle+\footnote{\href{http://squidle.org}{squidle.org}} is a powerful marine image data management platform for exploring and annotating underwater imagery.  With thousands of images and annotations, it provides a rich archive of underwater training data.  Squidle+ is used to source the training data for the base and novel classes of marine species from AUV images captured off the coastline of Tasmania.  Existing annotations of marine species are commonly in point format as this is the scientific approach often used to measure the presence and abundance of species by labelling randomly placed points on an image~\cite{Pavoni2019Challenges}.  The Segment Anything semantic segmentation model~\cite{Kirillov2023Segment} is used through the Squidle+ platform to provide a segmentation boundary around the object under the point annotation.  The boundary is then used to create bounding boxes for training and image masks for the copy-paste augmentation operation.

Our framework's aim is to train an object detector for a rare or novel class using images from base classes to pre-train the backbone of the object detector and augment the training data.  Existing point annotations are transformed efficiently into segmentation masks and bounding boxes for training.  Applying the framework to the example of handfish detection in AUV imagery demonstrated improved detection performance for objects with low annotations compared to training without the framework.
 
The contribution of this paper is a framework to improve detection performance for objects with low annotations by:
\begin{itemize}
\item using pre-training of the object detector backbone with annotated images of common base classes followed by fine-tuning with novel class data to create discriminative features
\item augmenting training data during fine-tuning by applying a copy-paste operation in two directions to address both low numbers of annotations and domain shift
\item efficiently generating bounding box annotations for object detection training using point annotations and Segment Anything segmentation model
\item demonstrating improved performance for one-stage and two-stage detectors by presenting results when applied to an image dataset of handfish and common marine species taken by two AUVs around the coastline of Tasmania
\end{itemize}
The datasets, including images, point annotations and segmentations, are publicly available on the Squidle+ marine imaging platform.

The remainder of this paper is organised as follows.  Section~\ref{Sec:RelatedWork} presents an overview of work related to the training of object detectors, with a particular focus on tools that accommodate low numbers of training examples.  Section~\ref{Sec:EndangeredSpecies} describes our framework and how it uses pre-training and copy-paste augmentation during training.  Section~\ref{Sec:Results} describes the application of the proposed framework to the detection of a rare species of fish captured in AUV imagery offshore of Tasmania and the results of the study, while Section~\ref{Sec:Conclusions} provides concluding remarks and directions for further study.

\section{Related Work}\label{Sec:RelatedWork}

\subsection{Object Detector models}
Deep learning object detectors can locate and classify an object after supervised training.  The main architectures for object detectors are one-stage and two-stage detectors.  Two-stage detectors such as Faster R-CNN~\cite{Ren2017FasterRCNN} have an initial stage for locating proposed objects for detection called a Region Proposal Network (RPN), followed by a second stage that classifies the object and refines the location.  A single-stage detector uses a single network to both locate and classify the object.  Examples of single-stage detectors include the series of You Only Look Once (YOLO) detectors~\cite{Bochkovskiy2020YOLOv4} and the Fully Convolutional One-Stage (FCOS)~\cite{Tian2019FCOS} object detector.  The architectures for both one- and two-stage detectors include a backbone for feature extraction and a final network layer to classify the detected object and locate the bounding box.

Training the detector is usually performed in a supervised manner with bounding boxes and classes.  However, generating bounding box annotations can be time-consuming~\cite{Chen2021PointsAsQueries,Lin2023Oysternet}.  Furthermore, there may be existing point annotation data available that may be useful for training classifiers.  This is particularly the case in the seafloor imaging context where a lot of historical effort has focused on generating point labels for classes of interest~\cite{Williams2019Leveraging}.  There has been some research into using point annotations to weakly supervise the training of the object detector~\cite{Chen2021PointsAsQueries,Ge2023}.  Our work uses a semantic segmentation model, Segment Anything~\cite{Kirillov2023Segment}, to efficiently generate broadly accurate bounding boxes using the existing point annotations.

\subsection{Few-shot learning}
Few-shot learning aims to train an object detector to detect novel classes with few examples by leveraging base classes with large numbers of annotations.  Kang et al. used a meta learner with feature reweighting by training with abundant base classes to create features that can generalise to novel classes~\cite{Kang2019FewShot}.  Pre-training of the detector's backbone on an adjacent task can improve object detection as shown with earlier object detector networks~\cite{Girshick2014RichFeature}.  More recently, a simple two-stage method uses a pre-training and fine-tuning step to further improve detection performance~\cite{Wang2020Frustratingly}.  Base class data is used for pre-training the object detector to provide a backbone that produces rich features for unseen novel classes.  Our framework is based on this approach.

\subsection{Data augmentation}
Data augmentation methods have been shown to improve object detection performance. In YOLOv4~\cite{Bochkovskiy2020YOLOv4}, the augmentations included photometric distortions adjusting brightness, contrast and colour and geometric distortions like random rotation and flipping.  Another successful augmentation operation is copy-paste where either the segmentation mask or the bounding box of the class instance is copied to another image.  Dvornik et al. used a method to merge the copied example into the background so it appeared seamless~\cite{Dvornik2018Modeling}.  Ghiasi et al. simplified the process for segmentation models by simply copying and pasting the segmented mask of the object without any blending with the background context~\cite{Ghiasi2021SimpleCopy}.  We use the same approach by copying positive examples to other images as well as copying negative examples to the novel class images.

\section{Detector Training with Few Labelled Examples}\label{Sec:EndangeredSpecies}
Our work introduces a framework that aims to train a detector with low numbers of positive training annotations.  This is undertaken in the particular context of detecting rare or unusual species but is applicable to a broader range of problems where low numbers of positive training examples are available.  We adopt terms from few-shot learning for our methodology.  The target species will be called the novel class, and the common marine species will be called the base classes.  Differing from other few-shot learning tasks, we aim to detect only one novel class to support monitoring the presence of an endangered species or other object of interest.  We also use fewer base classes compared to other work, with only 6 base classes compared to 20 or more in related work~\cite{Kang2019FewShot,Wang2020Frustratingly}.  The number of base classes was limited to those with available annotations but could be increased.

\begin{figure}[!htbp]
  \centering
  \includegraphics[width=0.46\textwidth]{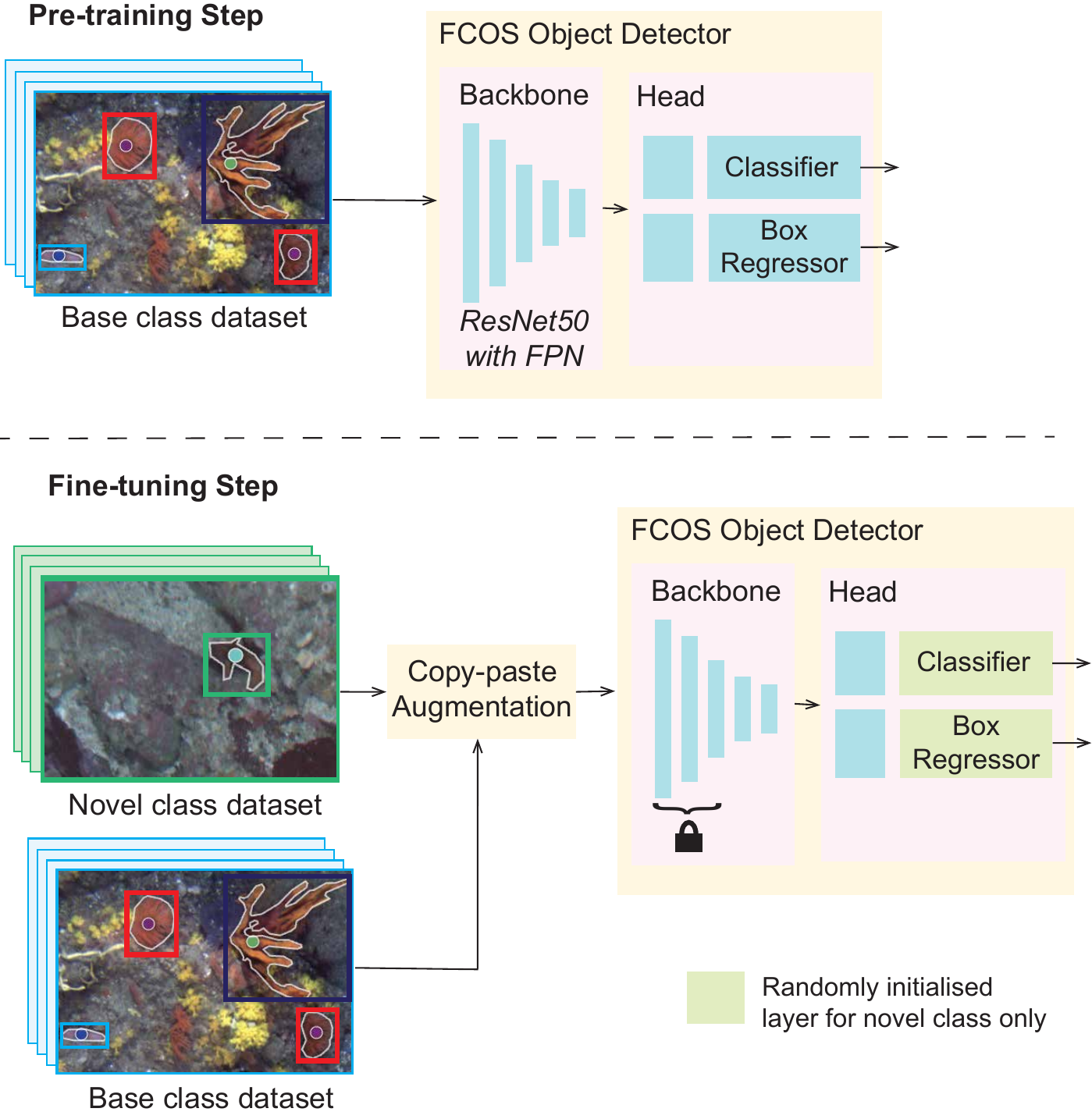}
  \caption{Detector training for one-stage detector, FCOS.  The pre-training step trains the object detector using a base class dataset.  In the fine-tuning step, training begins with the pre-trained detector with the box classifier and box regressor layer replaced with a newly initialised layer that only detects the novel class (boxes in green).  The first three layers of the backbone are frozen.  The base class dataset is used in the copy-paste operation to augment the dataset during fine-tuning.}
  \label{fig:pipeline}
\end{figure}

There are three main parts to the framework.  The first is preparing a dataset for training the object detector derived from images with point annotations of the novel and base classes. 
Second, an adaptation of the few-shot learning method by Wang et al. is used to pre-train the detector's backbone using the base class dataset~\cite{Wang2020Frustratingly}.  Finally, an augmentation method using copy-paste in two directions based on Ghiasi et al. is applied to increase the variety of images and reduce domain shift~\cite{Ghiasi2021SimpleCopy}.  Figure~\ref{fig:pipeline} provides an overview of the framework.  These three parts will be described in the next sections.

\subsection{Annotations and training data}
The training data includes one set of images with point annotations of the novel class and another with point annotations of the base class.  The annotations had already been collected for a statistical coverage survey using the methodology described in~\cite{Monk2018Evaluation}.  To generate bounding boxes for object detection training, the Segment Anything semantic segmentation model~\cite{Kirillov2023Segment} creates a segmentation mask for the object under the point annotation, which is then converted to a segmentation boundary.  The segmentation boundary is then used to create a bounding box for object detection training and to create a mask for copying the instance to another image.  

\begin{figure}[!t]
  \centering
  \includegraphics[width=0.39\textwidth]{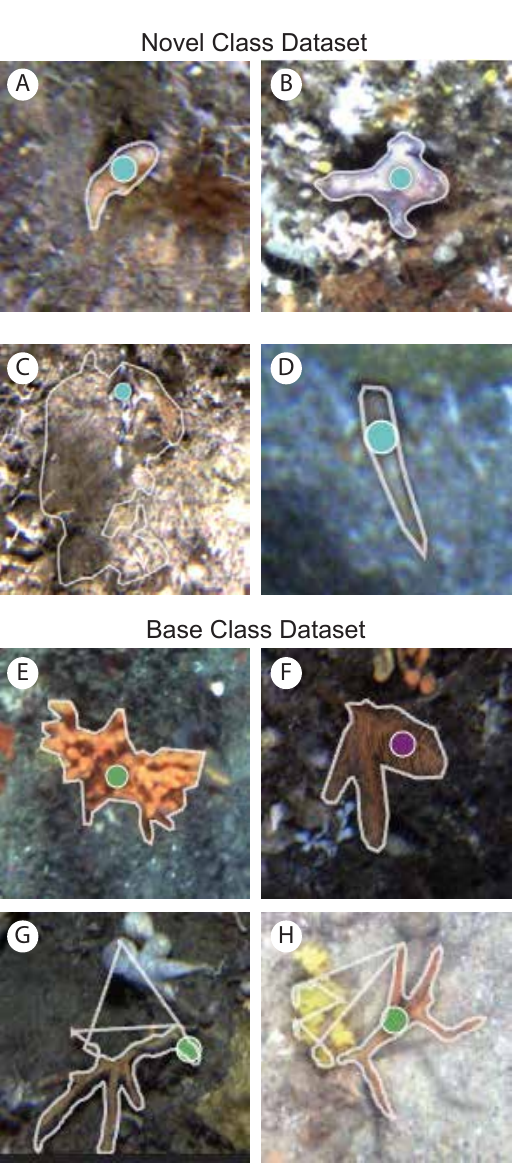}
  \caption{Examples of semantic segmentation boundaries generated from point annotations.  The first row for each dataset (A, B, E, F) show successful segmentations.  The row below (C, D, G, H) shows failed or poor-quality segmentation.  (C) incorrectly includes the seafloor around the handfish, while (D) removes the distinctive hand-like fins of the handfish.  Only errors in the handfish boundaries were manually corrected to ensure high-quality masks for the copy-paste operation.}
  \label{fig:seg_errors}
\end{figure}

Figure~\ref{fig:seg_errors} gives examples of successful and failed segmentation boundaries of both novel and base class instances.  Failed or poor quality boundaries occurred in around 1\% of base class annotations and around 15\% of handfish examples.  Errors in the handfish data were manually corrected to ensure a high-quality mask for the copy-paste operation.  The most common error was the exclusion of either the distinctive hand-like fins or the full tail.   No corrections were applied to the base class segmentation boundaries leading to a small amount of noisy training data for the pre-training step and the pasting of base class instances.

\subsection{Pre-training and Fine-Tuning}
Training starts with an initial pre-training step with the base class dataset, followed by fine-tuning with the novel class dataset.  
In pre-training, the backbone layers are all updated with the goal of generating features that adequately describe a range of marine species for classification.

In the fine-tuning step, the final layer of the object detection network is replaced with a randomly initialised classifier and box regression layer.  This replacement layer is only for the novel class and no longer detects the base classes.  Fine-tuning is performed with the novel class dataset, with 1/20 of the learning rate used during pre-training as in~\cite{Wang2020Frustratingly}. During this step, we freeze the layers of the first three levels of the backbone so the deeper layers remain fixed from the pre-training step, as shown in Figure~\ref{fig:pipeline}.

\begin{figure*}[ht!]
    \centering
    \includegraphics[width=0.85\textwidth]
    {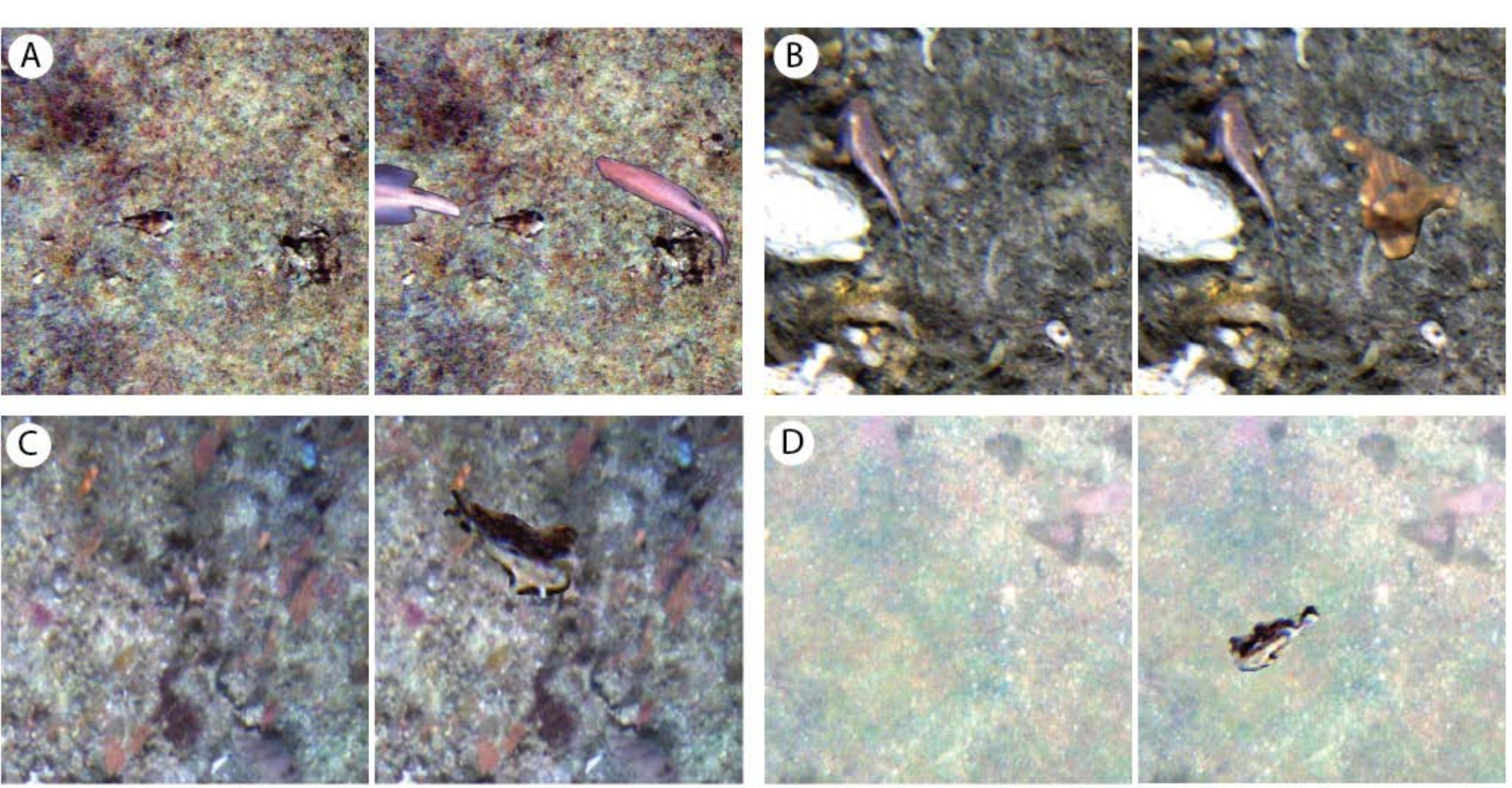}
    \caption{Examples of the two-way copy-paste augmentation operation.  Each pair of cropped images shows without (left) and with (right) copy-paste using the segmentation boundary as the mask.  The top row (A, B) contains images with handfish with a base class instance added.  The bottom row (C, D) is base class images with a handfish instance added.}
    \label{fig:copypaste}
\end{figure*}

\subsection{Copy-paste data augmentation}\label{method:aug}
Data augmentation compensates for the low number of annotations and domain shift. Basic horizontal and vertical flipping is applied to all images in pre-training and fine-tuning.    
The copy-paste operation is used as an additional augmentation operation to reduce domain shift by introducing images taken under different conditions to the novel class images.  The simple copy-paste method involves copying the selected segmentation masks to a new image~\cite{Ghiasi2021SimpleCopy}.  

First, an image from the novel and base class datasets is randomly selected.  One of these images is randomly cropped to provide more variety in the scaling of the image while ensuring that the instance of the novel class is kept within the cropping region. 
Next, the instances from one of the images are copied onto the other image.  Figure \ref{fig:copypaste} shows examples from this operation for handfish as the novel class and common marine species as the base classes.  While increasing the variation of images with the novel class, the copy-paste operation also increases the number of negative examples in the training data when copying the base classes to the novel class images.  The base classes include examples that could be false positives for the novel class.

\section{Experiments and Results}\label{Sec:Results}
We present results when applying the proposed framework to the problem of detecting an endangered marine species, handfish found off the coast of Tasmania.  Very few training examples of this species are available due to the low frequency of it being observed in the wild and the challenges of finding this cryptic species in the imagery of its natural habitat (see Figure~\ref{fig:handfish}).  

\subsection{Dataset}
All images and annotations were extracted from the Squidle+ marine image data management platform.  The novel and base class dataset images were taken by \emph{AUV Sirius} and \emph{AUV Nimbus} around Tasmania's coastline between 2009 and 2023. The datasets are publicly available on Squidle+\footnote{'ACFR Handfish Detection - IROS 2024', IMOS-UMI Squidle+, \href{http://squidle.org}{squidle.org}}.

For the handfish class, there are 284 images with a single handfish present in each image. A test set of the most recent images is used for evaluation containing 42 images with a single handfish captured between June 2021 to February 2023 by \emph{AUV Sirius} and \emph{AUV Nimbus}.  The remaining 242 images were randomly divided into training sets with 50, 100 and 200 sample sizes and a validation set with 42 images.

\begin{figure}[ht!]
  \centering
  \includegraphics[width=0.42\textwidth]{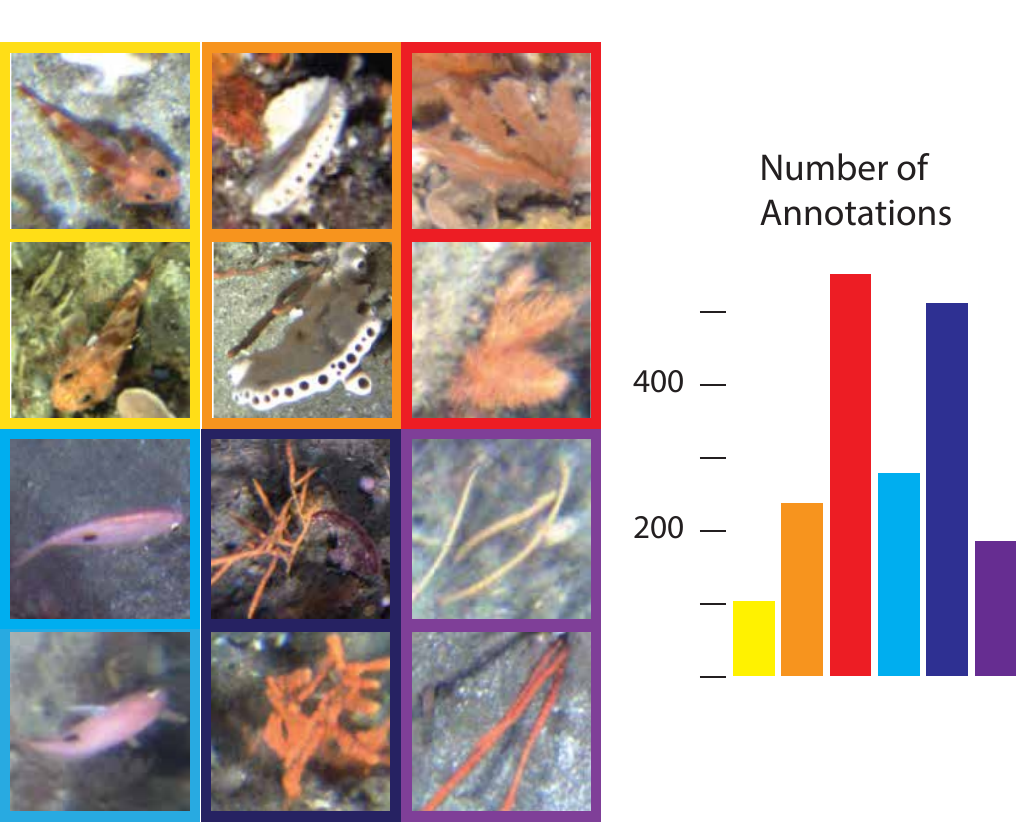}
  \caption{Examples of the six common marine species and the number of annotations used in the base class dataset.  The base class annotations are used in pre-training the backbone and for the two-way copy-paste augmentation operation.}
  \label{fig:base_classes}
\end{figure}

The base class dataset includes 275 images captured by \emph{AUV Sirius} and \emph{AUV Nimbus} with 1904 point annotations for six frequently occurring species.  These base classes were chosen because they had the highest number of existing annotations in Squidle+ and they also appear frequently in AUV images with handfish.  The images were annotated to capture all instances of the six species in the dataset.  There are between 100 and 550 annotations per class.  Examples of the species can be seen in Figure~\ref{fig:base_classes}, with each class distinguished by a colour border.

\begin{table*}[!ht]
\caption{Average Precision of Handfish detection with training sample sizes of 50, 100 and 200 for a two-stage and one-stage object detector. Average result and standard deviation from 5 runs are shown with the best values in red and second best in blue.\label{tab:combined_results}}
\centering
\begin{tabular}{c|cc|ccc|ccc}
Pre- & \multicolumn{2}{c|}{Copy-Paste Mask} & \multicolumn{3}{c|}{Two-Stage - Faster R-CNN} & \multicolumn{3}{c}{One-Stage - FCOS} \\
Train & Novel Class & Base Class & 50 & 100 & 200 & 50 & 100 & 200 \\
 \hline
N & None & None & $13.8\pm2.5$&\textcolor{blue}{$18.7\pm1.0$}&$19.1\pm1.9$ & $16.7\pm1.8$ & $19.5\pm1.7$ & $21.0\pm1.3$  \\
N & None & Segment. & $11.8\pm2.8$&$15.2\pm2.1$&$20.1\pm3.5$ & $18.2\pm3.1$ & $19.8\pm2.9$ & $22.8\pm3.8$  \\
N & Bounding Box & Segment. & $11.5\pm2.5$&$16.0\pm2.1$&$19.1\pm2.4$ & $17.2\pm4.1$ & \textcolor{blue}{$22.5\pm2.6$} & $21.6\pm3.8$  \\
N & Segment. & Segment. & $7.7\pm1.7$&$9.9\pm2.9$&{$18.0\pm2.0$} & $19.5\pm4.8$ & $17.5\pm3.8$ & \textcolor{blue}{$24.4\pm2.5$}  \\
 \hline
Y & None & None & \textcolor{blue}{$15.9\pm1.6$}&\textcolor{red}{$19.6\pm3.5$}&\textcolor{blue}{$21.1\pm3.6$} & $17.5\pm3.1$ & $18.2\pm2.7$ & $23.4\pm2.3$  \\
Y & None & Segment. & \textcolor{red}{$16.4\pm5.8$}&$17.4\pm3.8$&\textcolor{red}{$22.0\pm3.3$} & \textcolor{red}{$24.8\pm4.8$} & \textcolor{red}{$24.9\pm3.9$} & $21.0\pm3.2$  \\
Y & Bounding Box & Segment. & $10.3\pm4.2$&$13.7\pm3.4$&$20.3\pm3.0$ & \textcolor{blue}{$21.1\pm3.2$} & $16.5\pm4.5$ & $22.8\pm3.6$  \\
Y & Segment. & Segment. & $11.3\pm3.3$&$14.0\pm5.3$&$19.0\pm3.3$ & $18.7\pm6.8$ & $14.7\pm3.8$ & \textcolor{red}{$26.0\pm2.3$}  \\
\end{tabular}
\end{table*}

\subsection{Implementation Details} 
Similar to~\cite{Wang2020Frustratingly}, we use a Faster R-CNN object detector~\cite{Ren2017FasterRCNN} as the two-stage detector.  In addition, we apply our framework to a one-stage detector, FCOS~\cite{Tian2019FCOS}.  Both detectors use a ResNet-50 backbone with Feature Pyramid Network~\cite{Lin2017Feature} and we use the PyTorch implementation of the detector networks.  Also following~\cite{Wang2020Frustratingly}, pre-training uses a learning rate of 0.001 and all layers of the backbone are updated.  The fine-tuning step uses a learning rate of 0.0005, which is 1/20 of the pre-training learning rate.  During fine-tuning, the first three layers of the backbone are frozen so that only the higher-level features are updated.  Each training run is performed for 40 epochs with 1000 iterations per epoch and a batch size of 1. In all cases, training starts with a warmup phase for the first 1000 iterations, linearly increasing the learning rate from 1/1000 of its value to the learning rate, then decayed to 0.1 of its current value at 90\%, 95\% and 99.5\% of total iterations as in~\cite{Ghiasi2021SimpleCopy}.  All models were trained using an SGD optimiser with momentum of 0.9 and weight decay of 0.0005, and images were resized to 1000 pixels on the shorter side.  The training was performed on an NVIDIA A10G GPU.

\subsection{Results}
The framework is applied using novel datasets with a sample size of 50, 100 and 200 on a two-stage and one-stage detector.  The model from the final iteration of training is used for evaluation using Average Precision (AP) with an Intersection over Union (IoU) of 0.5.  Following other work on few-shot learning~\cite{Gao2022Decoupling}, the results are averaged over five training runs, each with a different random seed due to the small size of the training and test dataset.  

The results are shown in Table \ref{tab:combined_results} for training with and without the pre-training step and with variations on the copy-paste augmentation.  The first line of the table is used as the baseline case where no elements of the framework are applied.  In addition to results without any copy-paste augmentation, three variations of the copy-paste operation were evaluated with changes in whether the novel instances were used and, if they were, whether they used a bounding box mask or a segmentation boundary mask.  When the base class was used for copy-paste, the segmentation boundary was used.

\subsection{Discussion}
The framework's goal was to improve detector performance that utilised point annotations and a dataset of base class images and annotations for pre-training and data augmentation.  After being applied to a one-stage and two-stage detector with varying novel class sample sizes, the performance after pre-training alone increased in all but one case.  The pre-training step on more commonly available base class data appears to improve performance by extracting features that are discriminative for the unseen novel marine species.  The base class annotations were derived from the Segment Anything algorithm and included noisy data.  Despite sometimes inaccurate segmentations, the pre-trained backbone was able to improve performance (See Figure~\ref{fig:seg_errors}, (G) \& (H)).  
Greater performance gains from pre-training may be possible with improved segmentation boundaries and also increasing the number of base classes and annotations as used in Wang et al. \cite{Wang2020Frustratingly}.

Overall, the highest performance for 4 of the 6 cases was achieved using pre-training and one-way copy-paste of base class instances to novel class images.  This included the largest AP increase of 8.1 or 48\% for the one-stage detector with a novel sample size of 50.  The combination of the pre-trained backbone that could extract discriminative features for marine species as well as increased instances of negative examples from copy-paste of base class instances resulted in the highest performance.  Copy-paste was more beneficial for the one-stage detector, FCOS.  The framework is flexible enough to be applied to other detector architectures with a backbone and final classifier and box regression layer.  It could be successful if applied to other one-stage detectors.

\addtolength{\textheight}{-3cm}
   
\section{Conclusion}\label{Sec:Conclusions}
 
We have proposed a framework for detector training that increases performance for marine species that have low numbers of annotations.  
The overall framework is demonstrated on a dataset of endangered handfish producing increases in detection performance of up to 48\% with a sample size of 50.  Pre-training the detector with more commonly available marine species increased performance in most cases.  The copy-paste operation was most successful when used to add instances of a base class or negative examples to a novel class image during the fine-tuning step.  The framework has the flexibility to be applied to other one-stage detectors offering the opportunity to detect rare species in real-time on the AUV.  Future work could use the detector for adaptive planning to explore the local habitat of rare species when found.
Our framework provides a promising start to increasing the ability to efficiently identify threatened, endangered, and protected species from AUV images and was successfully used to identify previously unrecorded instances of handfish in AUV imagery, increasing the amount of data available to scientists interested in characterising the distribution of these critically endangered species.





\section*{ACKNOWLEDGMENT}
Images and annotations from \emph{AUV Sirius} and \emph{AUV Nimbus} were sourced from Australia’s Integrated Marine Observing System (IMOS) through the Squidle+ online platform.  IMOS is enabled by the National Collaborative Research Infrastructure Strategy (NCRIS), Australia.


\bibliographystyle{IEEEtran}
\bibliography{bibliography}

\begin{thebibliography}{10}
\providecommand{\url}[1]{#1}
\csname url@samestyle\endcsname
\providecommand{\newblock}{\relax}
\providecommand{\bibinfo}[2]{#2}
\providecommand{\BIBentrySTDinterwordspacing}{\spaceskip=0pt\relax}
\providecommand{\BIBentryALTinterwordstretchfactor}{4}
\providecommand{\BIBentryALTinterwordspacing}{\spaceskip=\fontdimen2\font plus
\BIBentryALTinterwordstretchfactor\fontdimen3\font minus \fontdimen4\font\relax}
\providecommand{\BIBforeignlanguage}[2]{{%
\expandafter\ifx\csname l@#1\endcsname\relax
\typeout{** WARNING: IEEEtran.bst: No hyphenation pattern has been}%
\typeout{** loaded for the language `#1'. Using the pattern for}%
\typeout{** the default language instead.}%
\else
\language=\csname l@#1\endcsname
\fi
#2}}
\providecommand{\BIBdecl}{\relax}
\BIBdecl

\bibitem{Stuart-Smith2020}
J.~Stuart-Smith, G.~Edgar, P.~Last, C.~Linardich, T.~Lynch, N.~Barrett, T.~Bessell, L.~Wong, and R.~Stuart-Smith, ``Conservation challenges for the most threatened family of marine bony fishes (handfishes: Brachionichthyidae),'' \emph{Biological Conservation}, vol. 252, p. 108831, 2020.

\bibitem{Perkins2022Changes}
N.~Perkins, J.~Monk, R.~Wong, S.~Willis, A.~Bastiaansen, and N.~Barrett, ``{Changes in rock lobster, demersal fish, and sessile benthic organisms in the Tasman Fracture Marine Park: comparisons between 2015 and 2021},'' Institute for Marine and Antarctic Studies, University of Tasmania, Tech. Rep., 2022.

\bibitem{Monk2018Evaluation}
J.~Monk, N.~S. Barrett, D.~Peel, E.~Lawrence, N.~A. Hill, V.~Lucieer, and K.~R. Hayes, ``{An evaluation of the error and uncertainty in epibenthos cover estimates from AUV images collected with an efficient, spatially-balanced design},'' \emph{PLoS ONE}, vol.~13, no.~9, 2018.

\bibitem{Perkins2021AnalysisOfATime}
N.~Perkins, J.~Monk, and N.~Barrett, ``{Analysis of a time-series of benthic imagery from the South-east Marine Parks Network},'' Institute of Marine and Antarctic Studies, Tech. Rep., 2021.

\bibitem{MassotCampos2023}
M.~Massot-Campos, F.~Bonin-Font, E.~Guerrero-Font, A.~Martorell-Torres, M.~M. Abadal, C.~Muntaner-Gonzalez, B.~M. Nordfeldt-Fiol, G.~Oliver-Codina, J.~Cappelletto, and B.~Thornton, ``Assessing benthic marine habitats colonized with posidonia oceanica using autonomous marine robots and deep learning: A eurofleets campaign,'' \emph{Estuarine, Coastal and Shelf Science}, vol. 291, 2023.

\bibitem{Kang2019FewShot}
B.~Kang, Z.~Liu, X.~Wang, F.~Yu, J.~Feng, and T.~Darrell, ``Few-shot object detection via feature reweighting,'' in \emph{Proceedings of the IEEE International Conference on Computer Vision}, vol. 2019-October, 2019, pp. 8419--8428.

\bibitem{Inoue2018CrossDomain}
N.~Inoue, R.~Furuta, T.~Yamasaki, and K.~Aizawa, ``{Cross-Domain Weakly-Supervised Object Detection Through Progressive Domain Adaptation},'' in \emph{Proceedings of the IEEE Computer Society Conference on Computer Vision and Pattern Recognition}, 2018, pp. 5001--5009.

\bibitem{Zhang2022C2FDA}
H.~Zhang, G.~Luo, J.~Li, and F.~Y. Wang, ``{C2FDA}: Coarse-to-fine domain adaptation for traffic object detection,'' \emph{IEEE Transactions on Intelligent Transportation Systems}, vol.~23, no.~8, pp. 12\,633--12\,647, 2022.

\bibitem{Munir2023}
M.~Munir, M.~Khan, M.~Sarfraz, and M.~Ali, ``{Domain Adaptive Object Detection via Balancing between Self-Training and Adversarial Learning},'' \emph{IEEE Transactions on Pattern Analysis and Machine Intelligence}, 2023.

\bibitem{Liu2020Towards}
H.~Liu, P.~Song, and R.~Ding, ``Towards domain generalization in underwater object detection,'' in \emph{Proceedings - International Conference on Image Processing, ICIP}, vol. 2020-October, 2020, pp. 1971--1975.

\bibitem{Er2023Research}
M.~J. Er, J.~Chen, Y.~Zhang, and W.~Gao, ``Research challenges, recent advances, and popular datasets in deep learning-based underwater marine object detection: A review,'' \emph{Sensors}, vol.~23, no.~4, 2023.

\bibitem{Bessell2022Prioritising}
T.~J. Bessell, J.~Stuart-Smith, N.~S. Barrett, T.~P. Lynch, G.~J. Edgar, S.~Ling, S.~A. Appleyard, K.~Gowlett-Holmes, M.~Green, C.~J. Hogg, S.~Talbot, J.~Valentine, and R.~D. Stuart-Smith, ``Prioritising conservation actions for extremely data-poor species: A risk assessment for one of the world's rarest marine fishes,'' \emph{Biological Conservation}, vol. 268, 2022.

\bibitem{Bessell2024Population}
T.~J. Bessell, R.~D. Stuart-Smith, O.~J. Johnson, N.~S. Barrett, T.~P. Lynch, A.~J. Trotter, and J.~Stuart-Smith, ``Population parameters and conservation implications for one of the world's rarest marine fishes, the red handfish (thymichthys politus),'' \emph{Journal of Fish Biology}, 2024.

\bibitem{Langenkamper2020}
D.~Langenkämper, R.~van Kevelaer, A.~Purser, and T.~W. Nattkemper, ``Gear-induced concept drift in marine images and its effect on deep learning classification,'' \emph{Frontiers in Marine Science}, vol.~7, 2020.

\bibitem{Wang2020Frustratingly}
X.~Wang, T.~Huang, T.~Darrell, J.~E. Gonzalez, and F.~Yu, ``Frustratingly simple few-shot object detection,'' in \emph{Proceedings of the 37th International Conference on Machine Learning}, vol. 119.\hskip 1em plus 0.5em minus 0.4em\relax JMLR.org, pp. 9919--9928.

\bibitem{Ghiasi2021SimpleCopy}
G.~Ghiasi, Y.~Cui, A.~Srinivas, R.~Qian, T.~Y. Lin, E.~D. Cubuk, Q.~V. Le, and B.~Zoph, ``Simple copy-paste is a strong data augmentation method for instance segmentation,'' in \emph{2021 IEEE/CVF Conference on Computer Vision and Pattern Recognition (CVPR)}, 2021, pp. 2917--2927.

\bibitem{Pavoni2019Challenges}
G.~Pavoni, M.~Corsini, N.~Pedersen, V.~Petrovic, and P.~Cignoni, ``Challenges in the deep learning-based semantic segmentation of benthic communities from ortho-images,'' \emph{Applied Geomatics}, vol.~13, no.~1, pp. 131--146, 2021.

\bibitem{Kirillov2023Segment}
A.~Kirillov, E.~Mintun, N.~Ravi, H.~Mao, C.~Rolland, L.~Gustafson, T.~Xiao, S.~Whitehead, A.~C. Berg, and W.-Y. Lo, ``{Segment Anything},'' in \emph{Proceedings of the IEEE/CVF International Conference on Computer Vision}, 2023, pp. 4015--4026.

\bibitem{Ren2017FasterRCNN}
S.~Ren, K.~He, R.~Girshick, and J.~Sun, ``{Faster R-CNN: Towards Real-Time Object Detection with Region Proposal Networks},'' \emph{IEEE Transactions on Pattern Analysis and Machine Intelligence}, vol.~39, no.~6, pp. 1137--1149, 2017.

\bibitem{Bochkovskiy2020YOLOv4}
A.~Bochkovskiy, C.-Y. Wang, and H.-Y.~M. Liao, ``{YOLOv4: Optimal speed and accuracy of object detection},'' \emph{arXiv preprint arXiv:2004.10934}, 2020.

\bibitem{Tian2019FCOS}
Z.~Tian, C.~Shen, H.~Chen, and T.~He, ``{FCOS}: Fully convolutional one-stage object detection,'' in \emph{Proceedings of the IEEE International Conference on Computer Vision}, vol. 2019-October, 2019, pp. 9626--9635.

\bibitem{Chen2021PointsAsQueries}
L.~Chen, T.~Yang, X.~Zhang, W.~Zhang, and J.~Sun, ``Points as queries: Weakly semi-supervised object detection by points,'' in \emph{Proceedings of the IEEE/CVF Conference on Computer Vision and Pattern Recognition}, 2021, pp. 8823--8832.

\bibitem{Lin2023Oysternet}
X.~Lin, N.~Sanket, N.~Karapetyan, and Y.~Aloimonos, ``{OysterNet: Enhanced oyster detection using simulation},'' in \emph{2023 IEEE International Conference on Robotics and Automation (ICRA)}.\hskip 1em plus 0.5em minus 0.4em\relax IEEE, 2023, pp. 5170--5176.

\bibitem{Williams2019Leveraging}
I.~D. Williams, C.~Couch, O.~Beijbom, T.~Oliver, B.~Vargas-Angel, B.~Schumacher, and R.~Brainard, ``Leveraging automated image analysis tools to transform our capacity to assess status and trends on coral reefs,'' \emph{Frontiers in Marine Science}, vol.~6, no. APR, 2019.

\bibitem{Ge2023}
Y.~Ge, Q.~Zhou, X.~Wang, C.~Shen, Z.~Wang, and H.~Li, ``Point-teaching: Weakly semi-supervised object detection with point annotations,'' in \emph{Proceedings of the 37th AAAI Conference on Artificial Intelligence, AAAI 2023}, vol.~37, 2023, pp. 667--675.

\bibitem{Girshick2014RichFeature}
R.~Girshick, J.~Donahue, T.~Darrell, and J.~Malik, ``Rich feature hierarchies for accurate object detection and semantic segmentation,'' in \emph{Proceedings of the IEEE Computer Society Conference on Computer Vision and Pattern Recognition}, 2014, pp. 580--587.

\bibitem{Dvornik2018Modeling}
N.~Dvornik, J.~Mairal, and C.~Schmid, ``Modeling visual context is key to augmenting object detection datasets,'' in \emph{Lecture Notes in Computer Science (including subseries Lecture Notes in Artificial Intelligence and Lecture Notes in Bioinformatics)}, vol. 11216 LNCS, 2018, pp. 375--391.

\bibitem{Lin2017Feature}
T.~Y. Lin, P.~Dollár, R.~Girshick, K.~He, B.~Hariharan, and S.~Belongie, ``Feature pyramid networks for object detection,'' in \emph{Proceedings - 30th IEEE Conference on Computer Vision and Pattern Recognition, CVPR 2017}, vol. 2017-January, 2017, pp. 936--944.

\bibitem{Gao2022Decoupling}
B.~B. Gao, X.~Chen, Z.~Huang, C.~Nie, J.~Liu, J.~Lai, G.~Jiang, X.~Wang, and C.~Wang, ``Decoupling classifier for boosting few-shot object detection and instance segmentation,'' in \emph{Advances in Neural Information Processing Systems}, vol.~35, 2022.

\end{thebibliography}

\end{document}